\documentclass[journal]{IEEEtran}

\usepackage{graphicx}
\usepackage{microtype}
\usepackage{amsmath,amssymb}
\usepackage{booktabs}
\usepackage{array}
\usepackage{url}
\usepackage{xcolor}

\usepackage{cite}
\usepackage{multibib}
\newcites{supp}{Supplementary References}
\usepackage[hidelinks]{hyperref}
\hypersetup{
  pdftitle={Extending Dynamic World Surface Water Mapping to Sentinel-1 with AlphaEarth Embeddings},
  pdfauthor={Rohit Mukherjee, Frederick Policelli, Beth Tellman, TC Chakraborty, Jonathan Giezendanner, Jonathan A. Sullivan, and Ning Sun}
}

\graphicspath{{figures/}}

\begin{document}

\title{Extending Dynamic World Surface Water Mapping to Sentinel-1 with AlphaEarth Embeddings}

\author{Rohit~Mukherjee, Frederick~Policelli, Beth~Tellman, TC~Chakraborty, Jonathan~Giezendanner, Jonathan~A.~Sullivan, and Ning~Sun%
\thanks{This work was supported in part by NASA under the ACCESS Program award 19-ACCESS19-0041 and in part by the U.S. Department of Energy, Office of Science, Biological and Environmental Research (BER), as part of the Integrated Coastal Modeling (ICoM) project (Grant KP1703110/75415), and by a DOE Early Career Award to TC Chakraborty and Pacific Northwest National Laboratory (PNNL). PNNL is operated by Battelle for the U.S. Department of Energy under Contract DE-AC05-76RL01830. This research used resources of the National Energy Research Scientific Computing Center (NERSC). The authors thank Google Earth Engine for imagery access and processing, and NASA JPL's OPERA project for sharing DSWx-S1 predictions.}%
\thanks{R. Mukherjee, TC Chakraborty, and N. Sun are with Pacific Northwest National Laboratory, Richland, WA 99354 USA (e-mail: rohit.mukherjee@pnnl.gov). F. Policelli is with NASA Goddard Space Flight Center, Greenbelt, MD 20771 USA. J. Giezendanner is with Fujitsu Research of America, Santa Clara, CA 95054 USA. B. Tellman and J. A. Sullivan are with the University of Wisconsin--Madison, Madison, WI 53706 USA. \emph{(Corresponding author: Rohit Mukherjee.)}}}

\markboth{}{}

\maketitle

\begin{abstract}
Dynamic World (DW) maps land use and land cover globally at 10\,m from Sentinel-2 (S2) imagery, but only for cloud-free observations, which limits where and when surface water can be mapped. We use the DW water class as weak supervision for a Sentinel-1 (S1) synthetic aperture radar (SAR) model so that DW-like water maps can be produced for every S1 acquisition. Google's AlphaEarth Foundations (AEF) annual embedding supplies spatial context, while S1 backscatter supplies the acquisition-time observation. On 53 globally distributed scenes with independent annotations of 3\,m PlanetScope imagery acquired within 48\,h of the S1 overpass, the S1-only model already reaches a pooled water intersection over union (IoU) of 0.77, comparable to 0.75 for the operational OPERA DSWx-S1 product, and adding AEF raises it to 0.85. The fused model improves on the S1-only model on 44 of 53 scenes and exceeds OPERA on 48, and on the independent S1S2-Water benchmark it reaches 0.94, compared with 0.87 for OPERA. Optical land-cover products can thus provide scalable training labels for SAR surface water mapping.
\end{abstract}


\section{Introduction}
\IEEEPARstart{F}{requent} surface water maps are needed to monitor river width and discharge \cite{allen2018grwl,durand2016discharge}, lake and reservoir storage \cite{yao2023lakes}, and flood exposure \cite{tellman2021floods}. Dynamic World (DW) classifies 10\,m Sentinel-2 (S2) images with less than 35\% cloud cover in near real time \cite{brown2022dynamicworld}, but clouds and cloud shadows limit its valid coverage, particularly in the tropics \cite{wilson2016cloud}. Long-term optical water records such as the Joint Research Centre (JRC) Global Surface Water dataset share this limitation \cite{pekel2016surfacewater}. Sentinel-1 (S1) C-band synthetic aperture radar (SAR) observes through clouds, day and night \cite{torres2012sentinel1}, and automated processing chains map water from its backscatter \cite{twele2016s1flood}, but smooth non-water surfaces, radar shadow, and other low-backscatter terrain can be confused with open water \cite{landuyt2019sarflood,jung2026dswxs1}. Operational SAR products therefore combine backscatter with static ancillary layers. OPERA DSWx-S1, for example, uses land cover, long-term water occurrence, and height above nearest drainage (HAND) to reduce such false detections \cite{jung2026dswxs1}. We ask whether a model can reproduce DW water maps from S1 on cloudy days, complementing S2-based DW to increase observation frequency.

To reduce false detections over radar-dark land, we add Google's AlphaEarth Foundations (AEF) embeddings as spatial context. AEF, one of several recent pixel-level embedding products \cite{tessera}, compresses a year of optical, radar, lidar, and other Earth observations into a 64-dimensional embedding for each 10\,m pixel \cite{brown2025alphaearth}, encoding persistent land cover and, indirectly, water occurrence. Unlike rule-based masks, AEF lets the model learn which spatial context helps interpret SAR observations.

Optical products have supplied weak supervision for water mapping. Sen1Floods11 used S2 spectral-index labels for S1 models \cite{bonafilia2020sen1floods11}, while DW labels have been used for fractional water mapping from 375\,m VIIRS imagery \cite{saunders2026viirs}. We use DW labels for acquisition-specific 10\,m SAR water mapping with annual AEF context. Other studies use AEF for U.S. post-event SAR flood segmentation \cite{beyondbackscatter} and benchmark foundation-model backbones for flood mapping \cite{geofm_flood}. Our focus is global surface water mapping with DW pseudo-labels rather than flood-event annotations.

This letter makes three contributions. First, we train an S1 water model on DW pseudo-labels alone, without manual annotation. Second, we isolate the contribution of annual AEF context with an S1-only baseline that shares the training data and schedule. Third, we evaluate both models against two references independent of DW, comprising 53 PlanetScope-based annotated scenes \cite{essd_gswd} and the S1S2-Water benchmark \cite{wieland2024s1s2water}, and compare them with the operational OPERA DSWx-S1 product \cite{opera2024dswxs1,jung2026dswxs1}.

\begin{figure*}[!t]
\centering
\makebox[0.800\textwidth][l]{\textbf{(a) Weak supervision for surface water mapping}}\\[3mm]
\includegraphics[width=0.800\textwidth]{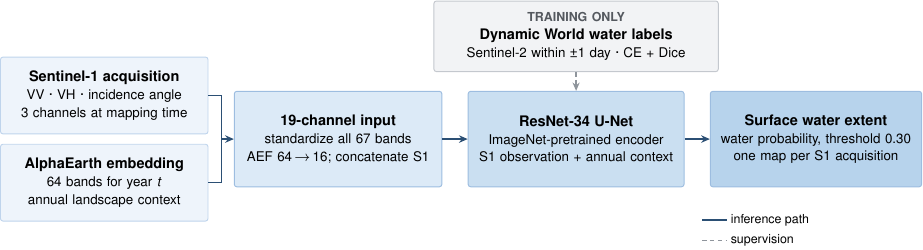}%
\\[3mm]
\makebox[0.800\textwidth][l]{\textbf{(b) Surface water predictions}}\\[1mm]
\includegraphics[width=0.800\textwidth]{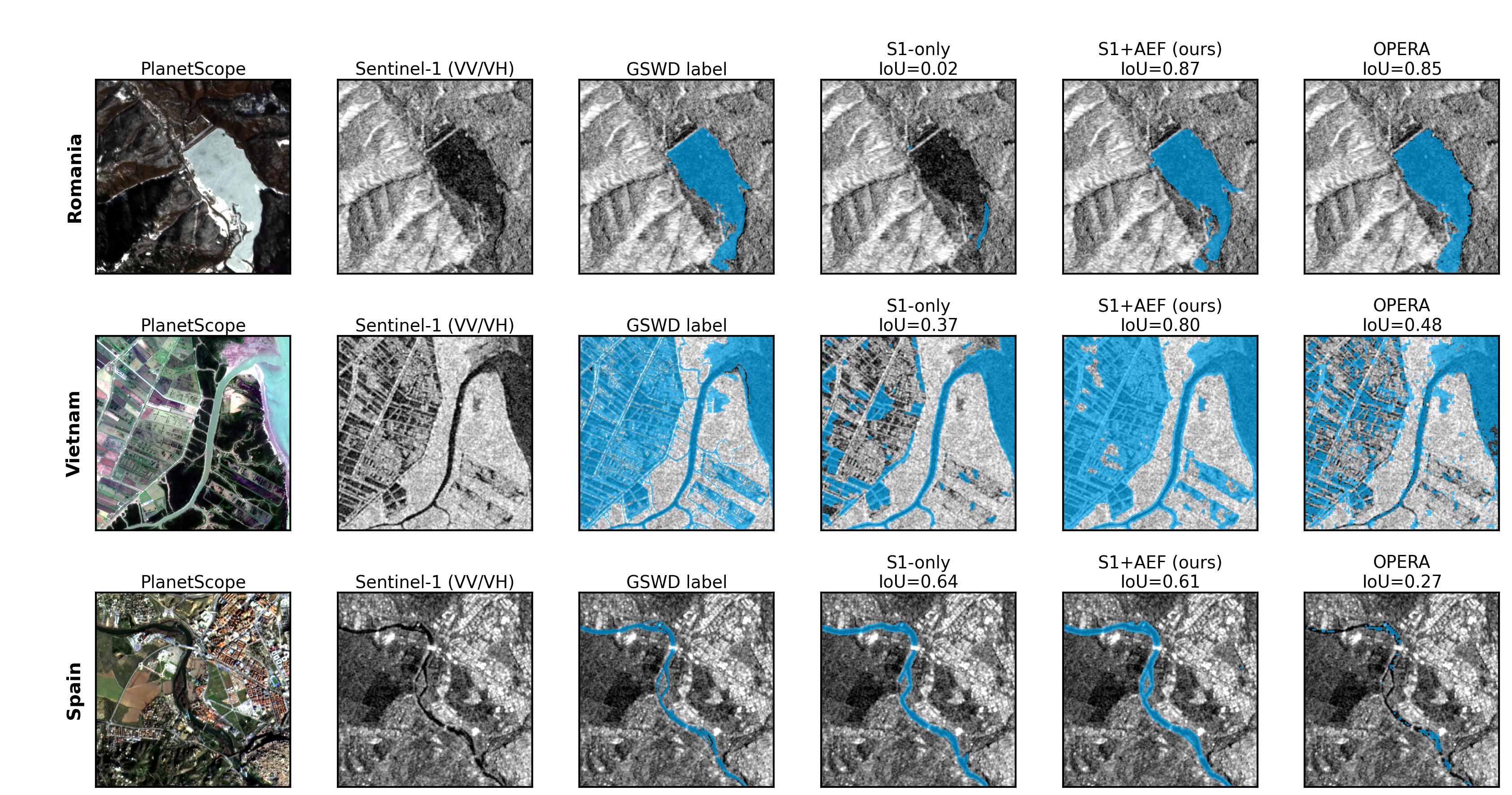}
\caption{(a) Weakly supervised S1 and AEF workflow; DW labels are used only in
training. (b) Three example scenes from the Global Surface Water Dataset (GSWD)
in Romania (SID23), Vietnam (SID46), and Spain (SID17) (10\,m inference,
threshold 0.30), with per-panel water intersection over union (IoU) against the
independent PlanetScope-based GSWD annotations on the common valid pixels. In
the Spain scene (SID17), a narrow river, the S1-only model is comparable to the
fused model. SID denotes the scene identifier in the PlanetScope reference
dataset.}
\label{fig:overview}
\end{figure*}

\section{Methods and Data}

\subsection{Weak Supervision from DW to S1}
Training sites were primarily sampled from the Surface Water and Ocean Topography (SWOT) River Database (SWORD) v16 river reaches \cite{altenau2021sword,altenau2023swordv16}. We randomly selected 20\% of river reaches on each continent and retained one S1--DW pair per sampled reach if acquisitions within 24\,h of each other were available. We kept only pairs with at least 90\% valid DW coverage, which left 4678 river tiles. Each tile covers approximately 10.7\,km per side, including surrounding land. We then added 600 tiles centered on lakes, seasonal water, reservoirs, and coastal sites using the same pairing and coverage criteria (Supplementary Section~S3). The label band of each paired DW image was converted to a binary mask (class~0, water $\rightarrow$ 1; all other classes $\rightarrow$ 0) and aligned pixel for pixel with the S1 tile; remaining no-data pixels were also set to 0. Labels therefore inherit DW classification errors \cite{venter2022global10m}, surface water changes between paired acquisitions, and cloud-masking errors. The final dataset has 5278 co-registered S1\,/\,AEF\,/\,DW triplets, split at random at the tile level (80/20, seeded) into 4222 training and 1056 validation tiles. For training, we sampled $512{\times}512$-pixel patches from these tiles. DW is used only as the training target; at inference the model receives S1 and AEF.

\subsection{S1 and Annual AEF Fusion}
We obtained S1 ground range detected (GRD) imagery from the Google Earth Engine \cite{gorelick2017google} \texttt{COPERNICUS/S1\_GRD} collection (Interferometric Wide swath, dual VV/VH polarization), which provides calibrated, orthorectified $\sigma^0$ backscatter in dB. We stacked VV, VH, and the incidence angle relative to the ellipsoid (\texttt{angle} band). The 64-band annual AEF embedding (v1, int8) was retrieved from the Source Cooperative archive \cite{sourcecooperativeaef} and reprojected to the S1 grid by nearest neighbor; it is constant within a calendar year and may include observations after the S1 acquisition. Each training tile uses the embedding of its acquisition year, except 98 tiles acquired in 2015--2016, before the first AEF year, which use the 2017 layer and therefore include later context.

S1 and AEF are fused at the input. A bias-free $1{\times}1$ convolution projects the 64 AEF bands to $k{=}16$ channels. These channels are concatenated with the three S1 bands to form a 19-channel input to a U-Net \cite{ronneberger2015unet} with an ImageNet-pretrained ResNet-34 encoder \cite{he2016resnet} (Fig.~\ref{fig:overview}(a)). The model is trained in four fine-tuning stages using a combined cross-entropy and Dice loss and the Ranger optimizer, with early stopping based on validation water IoU within each stage. Input bands are standardized using training-set statistics. Training details are provided in Supplementary Section~S3 and the code.

\subsection{Experimental Design}\label{sec:design}
We compare S1-only and S1+AEF($t$) models with the same training split, architecture, and schedule, differing only in the AEF input; $t$ denotes the S1 acquisition year. Both are trained with seeds 42--44; results use seed 42 unless otherwise stated. Additional analyses vary the projected AEF width and the training-set size (each over three seeds) and the embedding year (at inference and in training) (Section~\ref{sec:ablation}; Supplementary Section~S4).

\subsection{Independent Evaluation and Product Comparisons}\label{sec:eval}
The Global Surface Water Dataset (GSWD) \cite{essd_gswd} contains 100 manually annotated 3\,m PlanetScope-based maps spanning rivers, lakes, urban areas, and diverse biomes. We retained 53 with S1 GRD within 48\,h of PlanetScope (median 19.9\,h). S1 and AEF were exported at the 10\,m training resolution. Inference averages four-flip predictions on $512{\times}512$-pixel tiles with 64-pixel overlap; probabilities are bilinearly resampled to the 3\,m reference grid before thresholding (Supplementary Section~S2). Supplementary Section~S5 tests inference on inputs resampled to 3\,m.

GSWD and SWORD are independent sources, but training and evaluation sites were not geographically separated. Five evaluation scenes lie within 5\,km of a training-tile centroid and eight within 11\,km. Excluding these eight scenes leaves pooled S1+AEF water IoU essentially unchanged (0.855 for the remaining 45 versus 0.851 for all 53). The primary operating point is a fixed threshold of 0.30, selected from a DW-validation threshold sweep of an earlier model and retained to compare all reported models at the same operating point; it was not selected using GSWD or S1S2-Water reference labels. All methods are scored on the same pixels, those with reference values 0 or 1 and valid predictions from every model run and OPERA, comprising 93.4\% of the reference grid; scoring each method on its own valid pixels changes no pooled or per-scene mean IoU by more than 0.01. We report pooled precision, recall, and Dice (F1), and water IoU aggregated in two ways. Pooled metrics aggregate confusion counts across scenes, whereas per-scene IoU gives each scene equal weight. Bootstrap confidence intervals (CIs) quantify uncertainty in the per-scene mean, and paired Wilcoxon signed-rank tests evaluate within-scene differences between methods.

We compare primarily with OPERA DSWx-S1's 30\,m B02\_BWTR layer for the same S1 acquisition. Water includes inundated vegetation; HAND-excluded, layover/shadow, ocean, and no-data pixels are invalid. For overlapping OPERA product tiles (28 of 53 scenes), we union valid-pixel and water masks after nearest-neighbor resampling to 3\,m. Published OPERA validation scores only entirely water or non-water 30\,m pixels \cite{jung2026dswxs1}; we include all valid pixels, treating boundaries more strictly. Matched 30\,m scoring and pure-reference blocks test resolution and boundary sensitivity (Supplementary Section~S5). DW is assessed separately because no GSWD scene has a cloud-free DW map within $\pm$4\,days of PlanetScope (Section~\ref{sec:products}).

\subsection{S1S2-Water Benchmark}\label{sec:s1s2eval}
We evaluate 15 of 16 S1S2-Water test scenes \cite{wieland2024s1s2water} (Supplementary Table~S1), excluding one whose source S1 acquisition could not be reproduced (Supplementary Section~S1). These 100\,km $\times$ 100\,km scenes cover whole landscapes; water masks were manually refined by inspecting the S1 imagery. No benchmark data were used for training or threshold selection. OPERA was generated for the same acquisitions, and all methods are scored at 30\,m on common valid pixels (Supplementary Section~S1).

\section{Results}

\begin{table*}[!t]
\centering
\caption{Surface Water Mapping Accuracy on 53 Globally Sampled Surface Water Scenes (10\,m Inference Scored on the 3\,m Reference, Threshold 0.30, Common Valid Pixels; Model Seed 42)}
\label{tab:main}
\small
\begin{tabular}{llccccc}
\toprule
Method / product & Input & Precision & Recall & Dice & Water IoU (pooled) & Water IoU (per-scene mean [95\% CI]) \\
\midrule
S1-only              & S1                    & 0.883 & 0.854 & 0.869 & 0.768 & 0.617 [0.527, 0.702] \\
\textbf{S1\,+\,AEF ($t$)} & \textbf{S1 $+$ acq.-year AEF} & \textbf{0.891} & \textbf{0.950} & \textbf{0.920} & \textbf{0.851} & \textbf{0.741 [0.669, 0.806]} \\
\midrule
OPERA DSWx-S1        & \shortstack[l]{S1 $+$ static\\ancillary layers} & 0.859 & 0.851 & 0.855 & 0.747 & 0.591 [0.508, 0.670] \\
\bottomrule
\end{tabular}

\vspace{2pt}
{\footnotesize Intervals are bootstrap 95\% confidence intervals (CIs). Mean $\pm$ SD over seeds 42--44: pooled IoU 0.764\,$\pm$\,0.005 (S1-only), 0.850\,$\pm$\,0.001 (S1\,+\,AEF($t$)); per scene, Table~\ref{tab:ablation}.}
\end{table*}

\subsection{An S1-Only Model Trained on DW Labels}
Trained solely on DW water labels, S1-only reaches pooled IoU 0.768 on the independent PlanetScope reference, slightly above OPERA (0.747), and exceeds it on 36/53 scenes (Wilcoxon $p<0.05$). Agreement with DW-independent labels supports optical supervision of SAR water mapping.

\subsection{Annual AEF Context Improves the SAR Extension}\label{sec:aef}
\begin{figure}[!tb]
\centering
\includegraphics[width=\columnwidth]{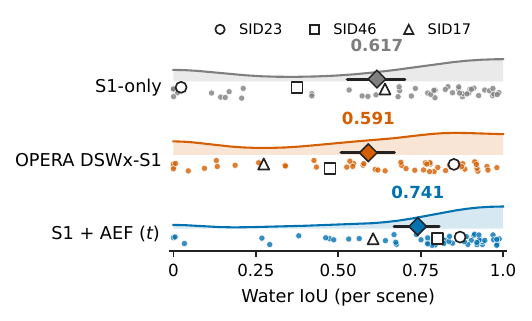}
\caption{Per-scene water IoU for 53 scenes (10\,m inference, threshold 0.30, common valid pixels). Curves show smoothed distributions, diamonds are means, and bars are bootstrap 95\% CIs. Outlined circles, squares, and triangles mark SID23, SID46, and SID17 from Fig.~\ref{fig:overview}(b). S1+AEF($t$) exceeds S1-only on 44/53 scenes and OPERA on 48/53 (Wilcoxon $p<0.05$).}
\label{fig:dist}
\end{figure}

AEF raises pooled water IoU from 0.768 to 0.851 (Table~\ref{tab:main}), a gain of $+0.083$ with the same training split and schedule. Recall rises from 0.854 to 0.950; precision changes little (0.883 to 0.891). Fusion improves on 44/53 scenes (Wilcoxon $p<0.05$), raising mean IoU from 0.62 (bootstrap 95\% CI 0.53--0.70) to 0.74 (0.67--0.81) (Fig.~\ref{fig:dist}). Gains are uneven, with a median of $+0.03$, and 13 scenes with gains above 0.2 account for 85\% of the summed gain; in six, S1-only misses most water (IoU below 0.2).

In WorldCover \cite{zanaga2022worldcover} overlays, 77\% of the false negatives corrected by AEF lie in WorldCover permanent water bodies, and 59\% of the corrected false positives lie on bare or sparsely vegetated land. AEF also adds false positives, 62\% of which lie in WorldCover permanent water bodies that the GSWD reference marks as dry, consistent with AEF encoding persistent water occurrence (Supplementary Section~S4).

\begin{table}[!t]
\centering
\caption{Sensitivity Analyses on the 53 Surface Water Scenes (10\,m Inference, Threshold 0.30, Common Valid Pixels; Mean $\pm$ SD over Three Seeds)}
\label{tab:ablation}
\footnotesize
\begin{tabular}{lcc}
\toprule
Configuration & Water IoU (per scene) & Recall \\
\midrule
\multicolumn{3}{l}{\emph{(a) AEF feature width $k$, full training set}}\\
S1-only ($k{=}0$) & 0.624\,$\pm$\,0.007 & 0.867\,$\pm$\,0.015 \\
$k{=}1$ & 0.730\,$\pm$\,0.006 & 0.940\,$\pm$\,0.006 \\
$k{=}2$ & 0.737\,$\pm$\,0.002 & 0.948\,$\pm$\,0.002 \\
$k{=}3$ & 0.741\,$\pm$\,0.004 & 0.950\,$\pm$\,0.003 \\
$k{=}4$ & 0.741\,$\pm$\,0.002 & 0.950\,$\pm$\,0.003 \\
\textbf{$k{=}8$} & \textbf{0.743\,$\pm$\,0.001} & \textbf{0.951\,$\pm$\,0.001} \\
$k{=}16$ & 0.741\,$\pm$\,0.001 & 0.951\,$\pm$\,0.000 \\
\midrule
\multicolumn{3}{l}{\emph{(b) Pseudo-labeled training tiles, width $k{=}16$}}\\
1000 & 0.666\,$\pm$\,0.011 & 0.899\,$\pm$\,0.011 \\
2000 & 0.740\,$\pm$\,0.001 & 0.951\,$\pm$\,0.004 \\
\textbf{3000} & \textbf{0.742\,$\pm$\,0.002} & \textbf{0.951\,$\pm$\,0.002} \\
4222 (all) & 0.741\,$\pm$\,0.001 & 0.951\,$\pm$\,0.000 \\
\bottomrule
\end{tabular}

\vspace{2pt}
{\footnotesize Bold: highest per-scene IoU in each panel. Recall is pooled over all pixels and scenes. SD is the sample SD over seeds 42--44, as elsewhere in the paper. Panels (a) and (b) share the full-split $k{=}16$ runs, whose seed 42 is the primary model of Table~\ref{tab:main}.}
\end{table}

\subsection{Sensitivity Analyses}\label{sec:ablation}
A single projected AEF channel, a learned combination of all 64 bands, captures 91\% of the per-scene IoU gain achieved at $k{=}16$, accuracy is similar from $k{=}3$ onward, and performance plateaus near 2000 training tiles in this experiment (Table~\ref{tab:ablation}). Replacing AEF($t$) with AEF($t{-}1$) at inference, with the trained weights fixed, lowers pooled IoU from 0.851 to 0.846, retaining 94\% of the gain over S1-only. A model trained and applied with AEF($t{-}1$) reaches 0.848, retaining 96\% (Supplementary Section~S4).

\subsection{Comparison with Existing Products}\label{sec:products}
Against OPERA DSWx-S1, the fused model raises pooled IoU by 0.104 and improves on 48/53 scenes (median $+0.10$, Wilcoxon $p<0.05$), with higher precision and recall (Table~\ref{tab:main}).

The advantage persists with matched resolution. Area-averaging probabilities and scoring both methods at 30\,m gives pooled GSWD IoU 0.855 for S1+AEF versus 0.749 for OPERA, with fusion exceeding OPERA on 47/53 scenes. On entirely water or non-water reference blocks, IoU is 0.916 versus 0.825. Across scoring variants, the IoU difference between fusion and OPERA changes by at most 0.013 (Supplementary Table~S2).

DW requires a nearby cloud-free observation. No GSWD scene has one within $\pm$4\,days of its reference, only two have one within $\pm$8\,days, and the median offset is 16\,days; unlike training sites, GSWD scenes were not screened for DW coverage. Nearest cloud-free DW maps yield pooled IoU ${\approx}0.60$, partly reflecting this mismatch.

\begin{table}[!t]
\centering
\caption{Accuracy on the 15 S1S2-Water Test Scenes (30\,m Grid, Threshold 0.30, Common Valid Pixels; Model Values Are Mean $\pm$ SD over Three Seeds)}
\label{tab:s1s2}
\footnotesize
\setlength{\tabcolsep}{2.2pt}
\begin{tabular}{lccccc}
\toprule
Method & Prec. & Recall & \shortstack{IoU\\(pooled)} & \shortstack{IoU\\(per scene)} & \shortstack{Scenes\\$>$\,OPERA} \\
\midrule
S1-only & 0.806 & 0.982 & 0.794\,$\pm$\,0.026 & 0.723\,$\pm$\,0.019 & 6--7 \\
\textbf{S1\,+\,AEF ($t$)} & \textbf{0.946} & \textbf{0.995} & \textbf{0.941\,$\pm$\,0.001} & \textbf{0.862\,$\pm$\,0.002} & \textbf{12} \\
\midrule
OPERA DSWx-S1 & 0.904 & 0.957 & 0.869 & 0.763 & -- \\
\bottomrule
\end{tabular}

\vspace{2pt}
{\footnotesize Precision and recall are pooled and averaged over seeds; the last column gives the range over seeds.}
\end{table}

\subsection{Independent S1S2-Water Benchmark}\label{sec:s1s2}
On S1S2-Water (Table~\ref{tab:s1s2}), fusion exceeds OPERA on 12/15 scenes for every seed (Wilcoxon $p<0.05$). S1-only does not differ consistently from OPERA ($p\ge0.56$); its errors are mainly false positives. On arid S1S2-Water scene~23 (0.8\% water; Supplementary Table~S1), it labels radar-dark terrain as water (IoU 0.03, mean over seeds), versus 0.81 for fusion and 0.83 for OPERA. OPERA exceeds fusion only on scenes~23, 35, and 82, by at most 0.04.

\section{Discussion and Conclusion}
DW pseudo-labels provide scalable optical supervision for S1 water mapping. Trained on them alone, the S1-only model matches or slightly exceeds OPERA DSWx-S1 on GSWD (0.77 versus 0.75), and annual AEF context improves accuracy on both GSWD and S1S2-Water. S1 supplies the acquisition-time observation; AEF is fixed within a calendar year and cannot by itself represent within-year changes in water extent. Its role is to supply spatial context for interpreting that observation.

\looseness=-1
The two benchmarks highlight different S1-only errors, and AEF reduces both. On GSWD, AEF mainly recovers missed water, most of it in WorldCover permanent water bodies (Section~\ref{sec:aef}); these overlays show where errors change, not why. On S1S2-Water, whose scenes cover whole landscapes with little water, S1-only errors are mainly false positives, and AEF raises precision from 0.81 to 0.95. The gain is concentrated in scenes where S1 alone fails.

\looseness=-1
In these experiments, annual context also makes the model less sensitive to training and input choices. Across three training seeds, per-scene IoU varies less for the fused model than for S1-only (SD 0.001 versus 0.007, Table~\ref{tab:ablation}). Using the S1 images distributed with S1S2-Water instead of our Earth Engine S1 lowers pooled S1-only IoU from 0.82 to 0.72 but fused IoU only from 0.94 to 0.93, and running the network on GSWD inputs resampled to 3\,m instead of the 10\,m training grid lowers S1-only IoU from 0.77 to 0.57, while the fused model stays at 0.85 (seed 42; Supplementary Section~S5). S1 preprocessing differs between processing chains, so this stability is useful in practice.

\looseness=-1
The fused model was neither trained nor tuned on S1S2-Water, yet at native resolution (threshold 0.50, as in \cite{wieland2024s1s2water}) its mean per-scene IoU (0.88) is close to the 0.85 reported for the best S1 model trained on that benchmark (Supplementary Section~S5). Scene sets and averaging differ, so these results indicate transfer without retraining.

\looseness=-1
OPERA DSWx-S1 reduces errors over radar-dark land with fixed rules on static layers, including WorldCover, JRC water occurrence, and HAND \cite{jung2026dswxs1,pekel2016surfacewater}; AEF serves a similar purpose but is learned jointly with S1 from the training labels. HAND could complement our model as an input, while learned embeddings could inform rule-based masks.

\looseness=-1
Several limitations remain. Annual context can favor areas of persistent water, where most false positives added by AEF lie (Section~\ref{sec:aef}). The model was not designed or evaluated for post-flood inundation. The model can also reproduce systematic DW label errors. The split is tile-level rather than basin-held-out, most training tiles (4678 of 5278) are centered on river reaches, and 10\,m predictions have limited accuracy on narrow channels (Supplementary Section~S5). Acquisition-year AEF can include observations after the S1 acquisition, so results with acquisition-year AEF are retrospective, not achievable in near real time. Prior-year AEF provides an alternative: using that layer, at inference or in training, retained 94--96\% of the pooled IoU gain (Section~\ref{sec:ablation}). The approach still depends on continued AEF availability.

\looseness=-1
The pseudo-label dataset can expand as DW observations accumulate, and the same weak-supervision strategy could be tested with other SAR sensors. Together with optical observations, acquisition-specific S1 maps could support more frequent monitoring of river width, lake and reservoir area, and shorelines.

\section*{Data and Code Availability}
Code, trained-model access, sample metadata, and instructions for
reconstructing the dataset from public inputs are available at
\href{https://github.com/Rohit18/surfaceWaterMappingGlobal/tree/v1.0}{\mbox{https://}\nolinkurl{github.com/Rohit18/surfaceWaterMappingGlobal}} (v1.0).

\bibliographystyle{IEEEtran}
\bibliography{refs}

\begin{thebibliography}{10}
\providecommand{\url}[1]{#1}
\csname url@samestyle\endcsname
\providecommand{\newblock}{\relax}
\providecommand{\bibinfo}[2]{#2}
\providecommand{\BIBentrySTDinterwordspacing}{\spaceskip=0pt\relax}
\providecommand{\BIBentryALTinterwordstretchfactor}{4}
\providecommand{\BIBentryALTinterwordspacing}{\spaceskip=\fontdimen2\font plus
\BIBentryALTinterwordstretchfactor\fontdimen3\font minus
  \fontdimen4\font\relax}
\providecommand{\BIBforeignlanguage}[2]{{%
\expandafter\ifx\csname l@#1\endcsname\relax
\typeout{** WARNING: IEEEtran.bst: No hyphenation pattern has been}%
\typeout{** loaded for the language `#1'. Using the pattern for}%
\typeout{** the default language instead.}%
\else
\language=\csname l@#1\endcsname
\fi
#2}}
\providecommand{\BIBdecl}{\relax}
\BIBdecl

\bibitem{allen2018grwl}
G.~H. Allen and T.~M. Pavelsky, ``Global extent of rivers and streams,''
  \emph{Science}, vol. 361, no. 6402, pp. 585--588, 2018.

\bibitem{durand2016discharge}
M.~Durand, C.~J. Gleason, P.-A. Garambois \emph{et~al.}, ``An intercomparison
  of remote sensing river discharge estimation algorithms from measurements of
  river height, width, and slope,'' \emph{Water Resour. Res.}, vol.~52, no.~6,
  pp. 4527--4549, 2016.

\bibitem{yao2023lakes}
F.~Yao, B.~Livneh, B.~Rajagopalan, J.~Wang \emph{et~al.}, ``Satellites reveal
  widespread decline in global lake water storage,'' \emph{Science}, vol. 380,
  no. 6646, pp. 743--749, 2023.

\bibitem{tellman2021floods}
B.~Tellman, J.~A. Sullivan, C.~Kuhn, A.~J. Kettner \emph{et~al.}, ``Satellite
  imaging reveals increased proportion of population exposed to floods,''
  \emph{Nature}, vol. 596, no. 7870, pp. 80--86, 2021.

\bibitem{brown2022dynamicworld}
C.~F. Brown, S.~P. Brumby, B.~Guzder-Williams \emph{et~al.}, ``Dynamic {World},
  near real-time global 10\,m land use land cover mapping with {Sentinel-2} and
  deep learning,'' \emph{Sci. Data}, vol.~9, p. 251, 2022.

\bibitem{wilson2016cloud}
A.~M. Wilson and W.~Jetz, ``Remotely sensed high-resolution global cloud
  dynamics for predicting ecosystem and biodiversity distributions,''
  \emph{PLoS Biology}, vol.~14, no.~3, p. e1002415, 2016.

\bibitem{pekel2016surfacewater}
J.-F. Pekel, A.~Cottam, N.~Gorelick, and A.~S. Belward, ``High-resolution
  mapping of global surface water and its long-term changes,'' \emph{Nature},
  vol. 540, pp. 418--422, 2016.

\bibitem{torres2012sentinel1}
R.~Torres, P.~Snoeij, D.~Geudtner \emph{et~al.}, ``{GMES Sentinel-1} mission,''
  \emph{Remote Sens. Environ.}, vol. 120, pp. 9--24, 2012.

\bibitem{twele2016s1flood}
A.~Twele, W.~Cao, S.~Plank, and S.~Martinis, ``{Sentinel-1}-based flood
  mapping: A fully automated processing chain,'' \emph{Int. J. Remote Sens.},
  vol.~37, no.~13, pp. 2990--3004, 2016.

\bibitem{landuyt2019sarflood}
L.~Landuyt, A.~Van~Wesemael, G.~J.-P. Schumann, R.~Hostache, N.~E.~C. Verhoest,
  and F.~M.~B. Van~Coillie, ``Flood mapping based on synthetic aperture radar:
  An assessment of established approaches,'' \emph{IEEE Trans. Geosci. Remote
  Sens.}, vol.~57, no.~2, pp. 722--739, 2019.

\bibitem{jung2026dswxs1}
J.~Jung, H.~Fattahi, S.~Jeong \emph{et~al.}, ``Towards global mapping of
  dynamic surface water extents using {Sentinel-1} {SAR} data,'' \emph{Remote
  Sens. Environ.}, vol. 337, p. 115326, 2026.

\bibitem{tessera}
Z.~Feng, C.~Atzberger, S.~Jaffer \emph{et~al.}, ``{TESSERA}: Temporal
  embeddings of surface spectra for earth representation and analysis,'' in
  \emph{Proc. IEEE/CVF Conf. Comput. Vis. Pattern Recognit. (CVPR)}, 2026, pp.
  34\,818--34\,831.

\bibitem{brown2025alphaearth}
\BIBentryALTinterwordspacing
C.~F. Brown, M.~R. Kazmierski, V.~J. Pasquarella \emph{et~al.}, ``{AlphaEarth
  Foundations}: An embedding field model for accurate and efficient global
  mapping from sparse label data,'' arXiv:2507.22291, 2025. [Online].
  Available: \url{https://arxiv.org/abs/2507.22291}
\BIBentrySTDinterwordspacing

\bibitem{bonafilia2020sen1floods11}
D.~Bonafilia, B.~Tellman, T.~Anderson, and E.~Issenberg, ``{Sen1Floods11}: A
  georeferenced dataset to train and test deep learning flood algorithms for
  {Sentinel-1},'' in \emph{Proc. IEEE/CVF Conf. Comput. Vis. Pattern Recognit.
  Workshops (CVPRW)}, 2020, pp. 210--211.

\bibitem{saunders2026viirs}
\BIBentryALTinterwordspacing
A.~Saunders, J.~Giezendanner, B.~Tellman, D.~Slayback, F.~Policelli, R.~Zhang,
  and R.~Mukherjee, ``Fractional water detection with {VIIRS} and deep learning
  to enhance global flood monitoring,'' {SSRN} preprint, 2026, under revision.
  [Online]. Available: \url{https://ssrn.com/abstract=6554000}
\BIBentrySTDinterwordspacing

\bibitem{beyondbackscatter}
\BIBentryALTinterwordspacing
S.~Thasma, Y.-H. Ho, and A.~Mostafavi, ``Beyond backscatter: {AlphaEarth}
  land-cover priors for rapid {SAR} flood segmentation across foundation
  backbones,'' arXiv:2606.29134, 2026. [Online]. Available:
  \url{https://arxiv.org/abs/2606.29134}
\BIBentrySTDinterwordspacing

\bibitem{geofm_flood}
S.~Kaushik, L.~Maurya, E.~Tellman, and Z.~Zhang, ``Assessing the value of
  geo-foundational models for flood inundation mapping: Benchmarking models for
  {Sentinel-1}, {Sentinel-2} and {PlanetScope} for end-users,'' \emph{IEEE J.
  Sel. Topics Appl. Earth Obs. Remote Sens.}, 2026.

\bibitem{essd_gswd}
R.~Mukherjee, F.~Policelli, R.~Wang, E.~Arellano-Thompson, B.~Tellman,
  P.~Sharma, Z.~Zhang, and J.~Giezendanner, ``A globally sampled
  high-resolution hand-labeled validation dataset for evaluating surface water
  extent maps,'' \emph{Earth Syst. Sci. Data}, vol.~16, pp. 4311--4323, 2024.

\bibitem{wieland2024s1s2water}
M.~Wieland, F.~Fichtner, S.~Martinis, S.~Groth, C.~Krullikowski, S.~Plank, and
  M.~Motagh, ``{S1S2-Water}: A global dataset for semantic segmentation of
  water bodies from {Sentinel-1} and {Sentinel-2} satellite images,''
  \emph{IEEE J. Sel. Topics Appl. Earth Obs. Remote Sens.}, vol.~17, pp.
  1084--1099, 2024.

\bibitem{opera2024dswxs1}
{OPERA}, ``{OPERA} dynamic surface water extent from {Sentinel-1} version 1,''
  NASA Physical Oceanography Distributed Active Archive Center (PO.DAAC), 2024,
  dataset.

\bibitem{altenau2021sword}
E.~H. Altenau, T.~M. Pavelsky, M.~T. Durand \emph{et~al.}, ``The surface water
  and ocean topography ({SWOT}) mission river database ({SWORD}): A global
  river network for satellite data products,'' \emph{Water Resour. Res.},
  vol.~57, no.~7, p. e2021WR030054, 2021.

\bibitem{altenau2023swordv16}
E.~H. Altenau, T.~M. Pavelsky, M.~T. Durand, X.~Yang, R.~P. d.~M. Frasson, and
  L.~Bendezu, ``{SWOT} river database ({SWORD}), version 16,'' Zenodo, 2023.

\bibitem{venter2022global10m}
Z.~S. Venter, D.~N. Barton, T.~Chakraborty, T.~Simensen, and G.~Singh, ``Global
  10 m land use land cover datasets: A comparison of {Dynamic World},
  {WorldCover} and {Esri Land Cover},'' \emph{Remote Sensing}, vol.~14, no.~16,
  p. 4101, 2022.

\bibitem{gorelick2017google}
N.~Gorelick, M.~Hancher, M.~Dixon, S.~Ilyushchenko, D.~Thau, and R.~Moore,
  ``{Google Earth Engine}: Planetary-scale geospatial analysis for everyone,''
  \emph{Remote Sensing of Environment}, vol. 202, pp. 18--27, 2017.

\bibitem{sourcecooperativeaef}
\BIBentryALTinterwordspacing
{Taylor Geospatial}, ``{AlphaEarth} foundations satellite embedding dataset,''
  Source Cooperative, 2025, dataset, v1. [Online]. Available:
  \url{https://source.coop/tge-labs/aef}
\BIBentrySTDinterwordspacing

\bibitem{ronneberger2015unet}
O.~Ronneberger, P.~Fischer, and T.~Brox, ``{U-Net}: Convolutional networks for
  biomedical image segmentation,'' in \emph{Medical Image Computing and
  Computer-Assisted Intervention (MICCAI)}, 2015, pp. 234--241.

\bibitem{he2016resnet}
K.~He, X.~Zhang, S.~Ren, and J.~Sun, ``Deep residual learning for image
  recognition,'' in \emph{Proc. IEEE Conf. Comput. Vis. Pattern Recognit.
  (CVPR)}, 2016, pp. 770--778.

\bibitem{zanaga2022worldcover}
D.~Zanaga, R.~Van De~Kerchove, D.~Daems \emph{et~al.}, ``{ESA WorldCover} 10 m
  2021 v200,'' Zenodo, 2022.

\end{thebibliography}


\begin{thebibliography}{1}
\providecommand{\url}[1]{#1}
\csname url@samestyle\endcsname
\providecommand{\newblock}{\relax}
\providecommand{\bibinfo}[2]{#2}
\providecommand{\BIBentrySTDinterwordspacing}{\spaceskip=0pt\relax}
\providecommand{\BIBentryALTinterwordstretchfactor}{4}
\providecommand{\BIBentryALTinterwordspacing}{\spaceskip=\fontdimen2\font plus
\BIBentryALTinterwordstretchfactor\fontdimen3\font minus
  \fontdimen4\font\relax}
\providecommand{\BIBforeignlanguage}[2]{{%
\expandafter\ifx\csname l@#1\endcsname\relax
\typeout{** WARNING: IEEEtran.bst: No hyphenation pattern has been}%
\typeout{** loaded for the language `#1'. Using the pattern for}%
\typeout{** the default language instead.}%
\else
\language=\csname l@#1\endcsname
\fi
#2}}
\providecommand{\BIBdecl}{\relax}
\BIBdecl

\bibitem{wieland2024s1s2waterdata}
M.~Wieland, F.~Fichtner, S.~Martinis, S.~Groth, C.~Krullikowski, S.~Plank, and
  M.~Motagh, ``{S1S2-Water}: A global dataset for semantic segmentation of
  water bodies from {Sentinel-1} and {Sentinel-2} satellite images (v1.0.1),''
  \url{https://zenodo.org/records/11278238}, 2024.

\bibitem{wang2025pld}
J.~Wang, C.~Pottier, C.~Cazals \emph{et~al.}, ``The {Surface Water and Ocean
  Topography} mission ({SWOT}) {Prior Lake Database} ({PLD}): Lake mask and
  operational auxiliaries,'' \emph{Water Resour. Res.}, vol.~61, no.~3, p.
  e2023WR036896, 2025.

\bibitem{lehner2024gdw}
B.~Lehner, P.~Beames, M.~Mulligan \emph{et~al.}, ``The {Global Dam Watch}
  database of river barrier and reservoir information for large-scale
  applications,'' \emph{Sci. Data}, vol.~11, p. 1069, 2024.

\bibitem{wessel1996gshhg}
P.~Wessel and W.~H.~F. Smith, ``A global, self-consistent, hierarchical,
  high-resolution shoreline database,'' \emph{J. Geophys. Res., Solid Earth},
  vol. 101, no.~B4, pp. 8741--8743, 1996.

\end{thebibliography}

\clearpage
\setcounter{table}{0}
\setcounter{figure}{0}
\renewcommand{\thetable}{S\arabic{table}}
\renewcommand{\thefigure}{S\arabic{figure}}
\renewcommand{\theHtable}{supp.\arabic{table}}
\renewcommand{\theHfigure}{supp.\arabic{figure}}
\makeatletter
\def\@IEEEsectpunct{\ }
\makeatother
\setcounter{topnumber}{3}
\markboth{}{}
\section*{Supplementary Material}

\section*{S1. S1S2-Water benchmark}
We use the Sentinel-1 reference masks for 15 of the 16 test scenes in S1S2-Water v1.0.1 \cite{wieland2024s1s2water}, \citesupp{wieland2024s1s2waterdata}. Scene 31 was excluded before scoring because its source S1 acquisition could not be reproduced from its metadata and image encoding. The masks were initialized from the S2 normalized difference water index (NDWI) and manually corrected against each sensor's image. S1S2-Water was assembled for general surface-water mapping rather than flood-event evaluation. ``Flooded'' in Table~\ref{tab:s1s2water} denotes a land-cover class, not a documented flood event. No benchmark data were used for training or threshold selection.

\paragraph*{Inputs and scoring.} Earth Engine S1 GRD VV, VH, and incidence angle were exported for the 22 source products underlying the 15 evaluated scenes, mosaicked in the order listed in the benchmark metadata, and paired with acquisition-year AEF on a 10\,m grid. All three seeds use four-flip augmentation. Scoring uses a 30\,m grid. Reference pixels are valid when at least 50\% of their area is valid, and water when at least 50\% of that valid area is water. Model probabilities are resampled bilinearly before thresholding; OPERA uses nearest neighbor. Common valid pixels cover 92.6--100\% of the reference-valid area per scene.

\paragraph*{OPERA processing.} Matching SLC acquisitions were processed with RTC-S1 v1.0.4 and DSWx-SAR v1.2, using Copernicus GLO-30 elevation, GLO-30 HAND, WorldCover 2020, and JRC water occurrence. Missing offshore elevation/HAND tiles were filled with zeros. We applied an upstream correction to the DSWx-SAR v1.2 edge-water exclusion to all scenes. B02\_BWTR class 1 is water; HAND exclusion, layover/shadow, ocean, and no-data classes are invalid.

\begin{figure*}[!t]
\centering
\includegraphics[width=0.72\textwidth]{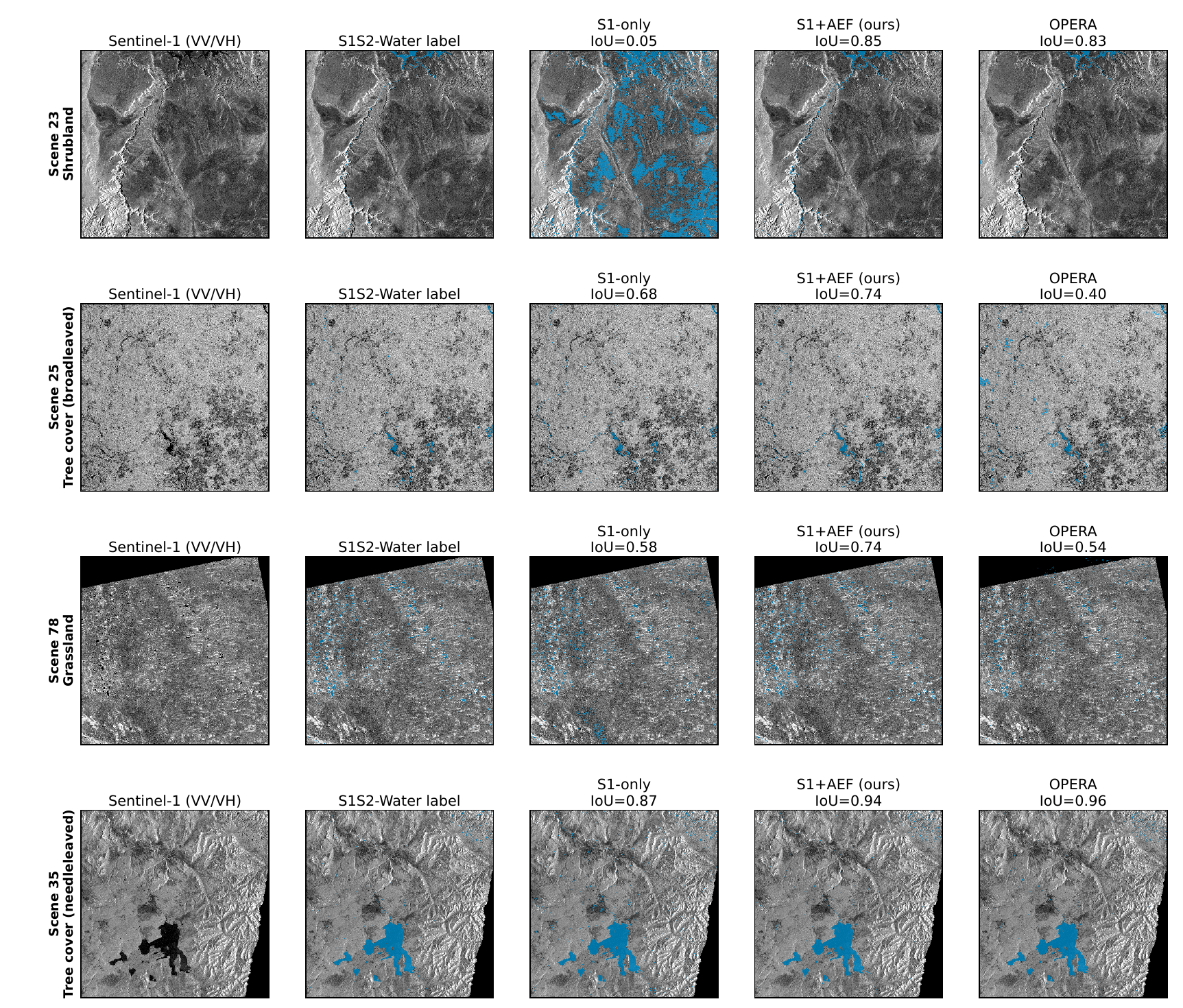}
\caption{Four S1S2-Water scenes showing S1 backscatter, reference water, and predictions from S1-only, S1+AEF, and OPERA. Models use seed 42 and threshold 0.50. IoU is scored at 30\,m on common valid pixels.}
\label{fig:s1s2water_examples}
\end{figure*}

\paragraph*{Results.} Table~\ref{tab:s1s2water} uses the fixed 0.30 threshold of the main letter. Fig.~\ref{fig:s1s2water_examples} and the values below use threshold 0.50, selected on DW validation. At 0.50, fused pooled IoU is $0.953\pm0.001$ versus $0.822\pm0.024$ for S1-only and 0.869 for OPERA. Fused predictions exceed OPERA on 13--14 scenes across seeds (Wilcoxon $p\le0.001$).

\begin{table*}[!t]
\centering
\caption{Per-scene water IoU on the 15 S1S2-Water test scenes (30\,m grid, common valid pixels). S1-only and S1+AEF: mean $\pm$ SD over three seeds, threshold 0.30 as in the main paper.}
\label{tab:s1s2water}
\footnotesize
\begin{tabular}{rllrrrr}
\toprule
Scene & S1 date & Land cover & Water frac. & OPERA & S1-only & S1+AEF \\
\midrule
23 & 2019-08-18 & Shrubland & 0.008 & 0.832 & 0.034$\pm$0.006 & 0.813$\pm$0.012 \\
25 & 2019-08-27 & Tree cover (broadleaved) & 0.007 & 0.400 & 0.630$\pm$0.005 & 0.671$\pm$0.016 \\
29 & 2019-09-02 & Herbaceous cover & 0.036 & 0.681 & 0.641$\pm$0.034 & 0.791$\pm$0.008 \\
33 & 2019-09-18 & Tree cover (broadleaved) & 0.071 & 0.804 & 0.800$\pm$0.014 & 0.884$\pm$0.006 \\
35 & 2019-09-16 & Tree cover (needleleaved) & 0.040 & 0.956 & 0.787$\pm$0.039 & 0.920$\pm$0.001 \\
47 & 2019-01-04 & Cropland rainfed & 0.081 & 0.788 & 0.858$\pm$0.002 & 0.875$\pm$0.003 \\
53 & 2019-04-16 & Herbaceous cover & 0.206 & 0.938 & 0.906$\pm$0.015 & 0.947$\pm$0.004 \\
57 & 2019-06-17 & Water bodies & 0.585 & 0.875 & 0.936$\pm$0.005 & 0.936$\pm$0.001 \\
75 & 2019-09-11 & Cropland rainfed & 0.087 & 0.587 & 0.862$\pm$0.013 & 0.849$\pm$0.003 \\
77 & 2019-09-18 & Tree cover (needleleaved) & 0.012 & 0.546 & 0.636$\pm$0.006 & 0.689$\pm$0.006 \\
78 & 2019-09-16 & Grassland & 0.020 & 0.541 & 0.328$\pm$0.114 & 0.705$\pm$0.014 \\
80 & 2019-10-14 & Shrub/herbaceous (flooded) & 0.144 & 0.622 & 0.857$\pm$0.009 & 0.923$\pm$0.006 \\
82 & 2019-09-26 & Tree cover (broadleaved) & 0.241 & 0.987 & 0.972$\pm$0.002 & 0.983$\pm$0.001 \\
88 & 2020-11-24 & Bare areas & 0.217 & 0.906 & 0.698$\pm$0.057 & 0.960$\pm$0.001 \\
89 & 2020-04-10 & Water bodies & 0.546 & 0.990 & 0.903$\pm$0.018 & 0.991$\pm$0.001 \\
\midrule
Pooled &  &  &  & 0.869 & 0.794$\pm$0.026 & 0.941$\pm$0.001 \\
Scene mean &  &  &  & 0.763 & 0.723$\pm$0.019 & 0.862$\pm$0.002 \\
\bottomrule
\end{tabular}
\end{table*}

\section*{S2. PlanetScope reference evaluation}
The 53 GSWD reference scenes are $1024\times1024$ pixels at 3\,m. Reference values 0 and 1 denote non-water and water, respectively, and are valid for scoring; any other value is excluded. S1--PlanetScope offsets have median 19.9\,h, maximum 44\,h, and are at most 24\,h for 30 scenes. Inference uses $512\times512$ tiles with 64-pixel overlap and four-flip augmentation; probabilities are resampled bilinearly to the reference grid before thresholding.

B02\_BWTR class 1 includes open water and inundated vegetation; classes 250, 251, 254, and 255 are invalid. After nearest-neighbor resampling, overlapping products are combined by a union of valid pixels and a union of water (28 scenes, up to eight products).

At threshold 0.50, pooled IoU is 0.86 for S1+AEF($t$), 0.85 for the prior-year substitution, and 0.77 for S1-only. Figure~2 in the main paper uses 10,000 scene-bootstrap replicates (seed 42) for mean-IoU confidence intervals; paired tests use Wilcoxon signed ranks.

\section*{S3. Training data and configuration}
\paragraph*{Sampling.} Additional sampling sites came from the SWOT Prior Lake Database \citesupp{wang2025pld}, JRC Global Surface Water \cite{pekel2016surfacewater}, Global Dam Watch \citesupp{lehner2024gdw}, and GSHHG \citesupp{wessel1996gshhg}. River tiles span October 2015--November 2025, with 95\% in 2020--2022; other tiles span 2020--2023. The tile-level split uses seed 42.

\paragraph*{Model and optimization.} The four fine-tuning stages comprise up to 2 frozen-encoder, 2 partially unfrozen, 10 fully unfrozen, and 10 tail epochs, with one-cycle learning rates of $10^{-4}$, $5\times10^{-7}$--$10^{-5}$, $6\times10^{-8}$--$3\times10^{-6}$, and $1.5\times10^{-7}$--$1.5\times10^{-6}$, respectively. The code provides the full training configuration.

\paragraph*{Computing and access.} Models were trained on NVIDIA A100 GPUs at the National Energy Research Scientific Computing Center (NERSC). Code, dependencies, and usage examples are provided at \href{https://github.com/Rohit18/surfaceWaterMappingGlobal/tree/v1.0}{\mbox{https://}\nolinkurl{github.com/Rohit18/surfaceWaterMappingGlobal}} (v1.0).

\section*{S4. Model sensitivity and error patterns}
\paragraph*{Embedding year.} Replacing acquisition-year AEF with prior-year AEF at inference, with the S1+AEF weights fixed (seed 42, threshold 0.30), gives pooled IoU 0.846 versus 0.851 and per-scene mean 0.732 [95\% CI 0.662, 0.797] versus 0.741 [0.669, 0.806], retaining 94\% of the pooled gain over S1-only (0.768). The per-scene decrease has mean 0.009 and median 0.004 ($p=0.004$); the substituted model exceeds S1-only on 41/53 scenes ($p<0.01$).

We also trained and applied the S1+AEF model ($k{=}16$, seed 42, same schedule) with prior-year AEF. Its training set omits the 137 tiles acquired in 2015--2017, which have no prior-year layer, and one tile with less than 90\% prior-year coverage (5140 of 5278 tiles). It reaches pooled IoU 0.848 and per-scene mean 0.734 [0.663, 0.800], retaining 96\% of the pooled gain. It exceeds S1-only on 41/53 scenes and OPERA on 48/53 (both $p<0.05$) and does not differ significantly from the acquisition-year model (mean per-scene difference 0.007, $p=0.07$).

\paragraph*{Error patterns.} Overlaid on WorldCover 2021 \cite{zanaga2022worldcover}, using pixels valid in both models, the false negatives corrected by AEF (1.38\,M pixels) lie mainly in permanent water bodies (77\%) and mangroves (13\%); the corrected false positives (0.84\,M) lie mainly on bare or sparse vegetation (59\%) and tree cover (20\%); and 62\% of the added false positives (0.81\,M) lie in permanent water bodies. AEF raises scene IoU by more than 0.05 on 24 scenes, changes it by at most 0.05 on 26, and lowers it by more than 0.05 on three. These associations describe where errors occur, not their cause.

\section*{S5. Input and scoring-resolution sensitivity}
\paragraph*{Inputs.} Values are pooled IoU at threshold 0.30 for seed 42. On GSWD, we applied the network to inputs resampled to the native 3\,m reference grid instead of the 10\,m training grid. S1-only falls from 0.768 to 0.567 (recall 0.854 to 0.586), whereas S1+AEF changes little (0.851 to 0.853). Narrow river channels remain difficult to map, as illustrated by SID26, where fused IoU is 0.268 at 10\,m versus 0.525 with 3\,m inputs. On S1S2-Water, replacing our Earth Engine S1 with the benchmark-provided S1 lowers S1-only from 0.820 to 0.722 and S1+AEF from 0.939 to 0.934 (common valid pixels of the model runs, without OPERA).

\paragraph*{Scoring resolution.} OPERA DSWx-S1 is a binary 30\,m product, whereas our models predict probabilities at 10\,m; on GSWD, the main paper resamples OPERA to 3\,m by nearest neighbor and model probabilities bilinearly. We rescored the same predictions in four ways (Table~\ref{tab:resolution}). These comprise (i) both at 30\,m, on $10\times10$ blocks of reference pixels, with area-averaged probabilities; blocks are valid if at least 50\% of their area is valid and water if at least 50\% of the valid area is water; (ii) the model thresholded on these blocks, like OPERA, and scored at 3\,m; (iii) as (i), only on blocks entirely water or entirely non-water in the reference, as in the OPERA validation \cite{jung2026dswxs1}; and (iv) OPERA's own grid, in the 46 scenes where all OPERA products share one grid. On S1S2-Water, model probabilities were area-averaged instead of resampled bilinearly. The S1+AEF--OPERA difference changes by at most 0.013. S1-only stays slightly above OPERA on GSWD at 30\,m (35/53 scenes; $p=0.019$ at 0.30, $p=0.057$ at 0.50).

\begin{table}[!t]
\centering
\caption{Resolution-Matched Comparison with OPERA DSWx-S1, Pooled IoU (Seed 42, Threshold 0.30, Common Valid Pixels of Each Variant)}
\label{tab:resolution}
\footnotesize
\setlength{\tabcolsep}{2pt}
\begin{tabular}{lccccc}
\toprule
Scoring & S1-only & S1+AEF & OPERA & \shortstack{S1+AEF\\$-$OPERA} & \shortstack{Scenes\\$>$\,OPERA}\\
\midrule
\multicolumn{6}{l}{\emph{GSWD}}\\
Main paper (3\,m) & 0.768 & 0.851 & 0.747 & 0.104 & 48/53\\
(i) Both at 30\,m & 0.767 & 0.855 & 0.749 & 0.106 & 47/53\\
(ii) Binary 30\,m model & 0.761 & 0.841 & 0.747 & 0.095 & 48/53\\
(iii) Pure 30\,m blocks & 0.837 & 0.916 & 0.825 & 0.091 & 47/53\\
(iv) OPERA grid$^{a}$ & 0.764 & 0.854 & 0.751 & 0.103 & 42/46\\
\midrule
\multicolumn{6}{l}{\emph{S1S2-Water (30\,m)}}\\
Main paper (bilinear) & 0.824 & 0.940 & 0.869 & 0.071 & 12/15\\
Area-averaged & 0.825 & 0.941 & 0.869 & 0.073 & 12/15\\
\bottomrule
\end{tabular}

\vspace{2pt}
{\footnotesize $^{a}$\,46 scenes; variant (i) gives a difference of 0.106 on these scenes. S1S2-Water values use seed 42; main Table~III averages three seeds.}
\end{table}

\paragraph*{Native resolution.} At the native S1S2-Water resolution and threshold 0.50, mean per-scene IoU is $0.748\pm0.018$ (S1-only) and $0.876\pm0.002$ (fused), compared with 0.850 for the best published S1 model \cite{wieland2024s1s2water}. The published evaluation includes scene 31 and uses weighted scene averages, whereas we give each of the 15 evaluated scenes equal weight.

\bibliographystylesupp{IEEEtran}
\bibliographysupp{refs}

\end{document}